\documentclass{article}

\PassOptionsToPackage{numbers, square, sort&compress}{natbib}

    \usepackage[preprint]{neurips_2021}

\usepackage{wrapfig}
\usepackage[utf8]{inputenc} 
\usepackage[T1]{fontenc}    
\usepackage{hyperref}       
\usepackage{url}            
\usepackage{booktabs}       
\usepackage{amsfonts}       
\usepackage{nicefrac}       
\usepackage{microtype}      
\usepackage{xcolor}         
\usepackage{amsmath}
\usepackage{graphicx}
\usepackage{textcomp}

\title{VA-GCN: A Vector Attention Graph Convolution Network for learning on Point Clouds}  

\author{
    Haotian Hu \\
    College of Optical Science and Engineering \\
    Zhejiang University \\
    \texttt{hht1996ok@zju.edu.cn} \\
    \And
    Fanyi Wang\thanks{Corresponding Author.} \\
    College of Optical Science and Engineering \\
    Zhejiang University \\
    \texttt{11730038@zju.edu.cn} \\
    \And
     Huixiao Le \\
    Peking University\\
    \texttt{interesting@pku.edu.cn} \\
}

\begin{document}

\maketitle

\begin{abstract}
Owing to the development of research on local aggregation operators, dramatic breakthrough has been made in point cloud analysis models. However, existing local aggregation operators in the current literature fail to attach decent importance to the local information of the point cloud, which limits the power of the models. To fit this gap, we propose an efficient Vector Attention Convolution module (VAConv), which utilizes K-Nearest Neighbor (KNN) to extract the neighbor points of each input point, and then uses the elevation and azimuth relationship of the vectors between the center point and its neighbors to construct an attention weight matrix for edge features. Afterwards, the VAConv adopts a dual-channel structure to fuse weighted edge features and global features. To verify the efficiency of the VAConv, we connect the VAConvs with different receptive fields in parallel to obtain a Multi-scale graph convolutional network, VA-GCN. The proposed VA-GCN achieves state-of-the-art performance on standard benchmarks including ModelNet40, S3DIS and ShapeNet. Remarkably, on the ModelNet40 dataset for 3D classification, VA-GCN increased by 2.4\% compared to the baseline. Codes and pre-trained models are available at \url{https://github.com/hht1996ok/VA-GCN}.
\end{abstract}
\section{Introduction}

With the booming of 3D vision in fields such as autonomous vehicles and robotics, 3D point cloud has become an emerging important research topic. In recent years, machine learning and computer vision have made a significant breakthrough in 3D point cloud processing, greatly improving the performance of point clouds in various tasks (3D shape classification [1,2,3,4], 3D semantic segmentation [1,2,3,7,8], 3D object detection [5,6], etc.). Due to the disorder, sparseness and irregularity of the point clouds, how to effectively use the point cloud information and capture the relationship between points comes to be the focus of this field.

\begin{figure}[htbp]
    \centering
    \includegraphics[width=\linewidth]{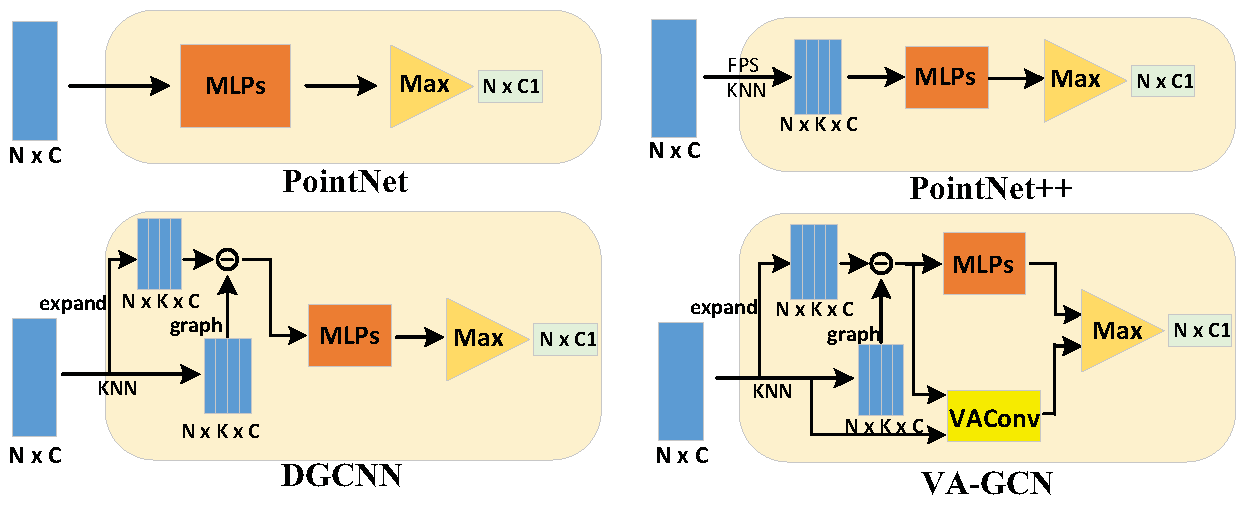}
    \caption{The schematic diagram of VA-GCN is compared with previous works. NxC represents the input size, and NxC1 represents the output size. Different from the previous networks, VA-GCN uses a dual-channel structure, which combines high-dimensional relative structure with low-dimensional relative structure.}
    \label{fig0}
\end{figure}

As is shown in Figure \ref{fig0}, PointNet [1] is a pioneering work that uses Multi-Layer Perceptron (MLP) to independently learn point features and uses a max-pooling layer to aggregate individual point features into a global representation. However, PointNet neglects the local information of point clouds. As such, Qi et al. proposed PointNet++ [2], which utilizes the local information and hierarchically processes a set of points, but pointNet++ ignores the geometric topology information between points. DGCNN [9] attempts to use the point cloud geometric information, which utilizes KNN to select the

nearest k neighbor points of each input point, builds a directed graph to extract geometric features, and extracts the edge features between the center point and its neighbor points through EdgeConv [9]. Unfortunately, DGCNN does not consider the direction of the vector, leading to the loss of useful information. Geo-CNN [26] aggregates local information according to the angle between the relative vector and the three coordinate axes, and establishes a standard basis for six directions to simulate convolution operations. However, Geo-CNN does not pay attention to the different influences of neighbor points around the central point in the process of information aggregation, resulting in the addition of irrelevant edge features. 

To solve the problems above, we propose a new dual-channel local information aggregation module named VAConv to assign greater weights to more important features in edge features. The advantage of VAConv is that it not only has explicit modelling of the local information aggregation process but also can flexibly change the size of the receptive field to make full use of the input information. What's more, by stacking the VAConv modules with different receptive fields, we constructed a Vector Attention Graph Convolution Network(VA-GCN), which hierarchically extracts the local information of the point clouds and fuses features with different scales. The proposed VA-GCN achieves state-of-the-art performance on standard benchmarks including ModelNet40 [8] (for 3D classification), S3DIS [40] (for 3D semantic segmentation) and ShapeNet [35] (for 3D part segmentation). Remarkably. for the classification task, the overall accuracy of VA-GCN is 2.1\% higher than baseline, which to the best of our knowledge, is the highest score so far. Our contributions are summarized as follows:

\begin{itemize}
    \item We propose an efficient Vector Attention Convolution module (VAConv), which uses a geometric representation of point clouds to explicitly construct the weight matrix of the local edge features. Meanwhile, global information constrained by relative vectors are added into local information to enrich the semantics of output features.
    \item By stacking EdgeConv and VAConv, we construct a dual-channel multi-scale local information aggregation model VA-GCN, which adds low-dimensional and high-dimensional relative geometric relations to the global semantics.
    \item We conduct extensive experiments on VA-GCN, and the results indicate that VA-GCN can achieve state-of-the-art performance on three benchmark datasets, for 3D classification, 3D semantic segmentation and 3D part segmentation respectively. We creatively propose Multi-Sample Inference(MSI), and after performing MSI, the highest score has been achieved on the ModelNet40 benchmark. 
\end{itemize}
\section{Related works}
Existing methods in literature can be roughly divided into three categories. Firstly, Multi-view methods [10,11,12,13] project point clouds into multiple two-dimensional views and use convolution to extract the features of each view, then perform feature fusion. Secondly, volumetric-based methods [14,8,16,17] voxelizes point clouds into a 3D grid, and uses 3D convolution to extract features from adjacent grids. Thirdly, point-based methods [1,2,3,4,18,19,20] is based on point clouds, which uses shared MLP to independently model each point, and then utilizes a symmetric function to aggregate the global features of point clouds.
\begin{figure}[htbp]
    \centering
    \includegraphics[width=\linewidth]{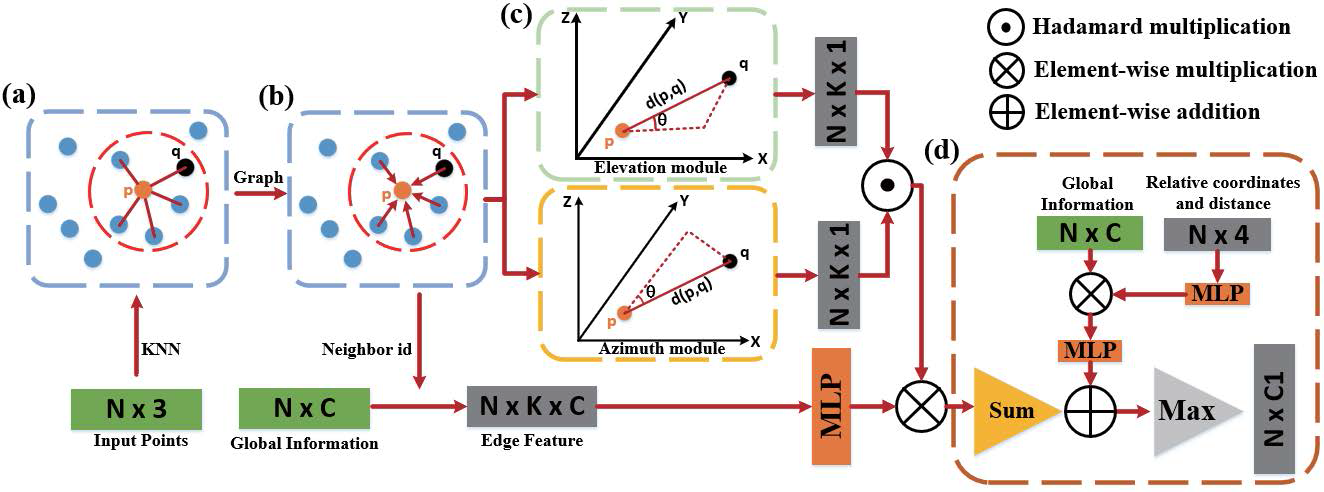}
    \caption{VAConv schematic diagram. N represents the number of points, and K represents the number of selected neighbor points. The input (green box) is the absolute position (with size of Nx3) and the global information (with size of NxC), \textbf{(a)} the KNN algorithm is used to get the neighbor points of each point. \textbf{(b)} Establish a directed graph to represent the adjacency of the center point, from which edge features and relative vectors could be obtained.  \textbf{(c)} Shows the calculation process of elevation and azimuth. \textbf{(d)} Represents the aggregation process of edge features and global features. Relative coordinates and distance (Nx4) is the output of \textbf{(b)}. }
    \label{fig1}
\end{figure}
\subsection{Multi-view methods}

Early works projected unstructured point clouds into multiple two-dimensional views, used 2D convolution to extract information from different views, and fused information from these sources to classify the point cloud accurately. The major challenge of this approach is how to fuse multi-view information. MVCNN [11] is a pioneering work, which uses a max-pooling to aggregate multi-view information into global information. However, the max-pooling only retains the largest element, which will inevitably cause information loss. Aimed at this problem, Wei et al. proposed View-GCN [21] which uses a directed graph and treats each view as a graph node. Max-pooling is performed on the graph nodes of all levels to obtain the global shape descriptors. Multi-view methods can directly use two-dimensional convolution after projection, which is convenient to implement, but are too slow to be suitable for general scenes.

\subsection{Volumetric-based methods}

The volumetric-based methods divide point clouds into a uniform spatial 3D grid, and then use a 3D convolutional neural network for 3D shape classification. Maturana et al. [14] introduced VoxNet to implement robust 3D target detection. Wu et al. [8] proposed 3D shapeNets, which uses DBN convolutional networks to learn the distribution of points with different three-dimensional shapes. Although those methods have achieved decent results, time and memory usage will explode with the increase of resolution. In order to solve this problem, recent researches have proposed sparse representation to reduce the demand for memory. OctNet [16] uses shallow octrees to represent the scene along the regular grid, and uses bit string representation to improve coding efficiency. Compared with the dense input network model, OctNet remits quite an amount of computational resources consumption but is still computational-intensive.

\subsection{Point-based methods}

The proposal of PointNet [1] paved the way for researchers. Based on the permutation invariance of point clouds, PointNet uses MLP to extract global information and aggregate features through the max-pooling layer. Although impressive results have been achieved, PointNet only processes each point individually and does not use the local information of point clouds, which will inevitably lead to the loss of key information. The successor, PointNet++ [2], introduces the concept of local information into the 3D point cloud analysis model. PointNet++ uses MLP to aggregate fine local information of neighbor points layer by layer, and obtains global descriptors through the local information obtained by max-pooling layer aggregation. 

Subsequent 3D point cloud analysis works mainly focus on the research of point cloud local information aggregation. PATs [28] uses the relative positions and absolute positions of points to represent each point and then extracts the local information of the point cloud. DGCNN [9] establishes a directed graph between the center point and its neighbor points, uses EdgeConv as a feature extraction function to extract the features of each edge, and aggregates local information through the max-pooling layer. Yan et al. [22] uses the Adaptive Sampling(AS) module to adaptively adjust the FPS algorithm. In order to solve the problems of long-running time and large memory requirements for large-scale point clouds, Xu et al. proposed Grid-GCN [20]. Furthermore, Grid-GCN proposed a fast sampling method based on Voxel, which combines volumetric-based methods and point-based methods, shortening the sampling time by 5 times. 

At the same time, a large number of models that simulate convolution operations to obtain global descriptors by carefully designing convolution kernels have emerged. RS-CNN [4] maps low-dimensional relations such as relative position and distance to high-level relations to simulate the convolution operation. The edge features along each direction of the base in Geo-CNN [26] are independently weighted by a learnable matrix related to one direction, then local information is aggregated based on the angle between the relative vector and the three coordinate axes. A-CNN [27] proposed an annular convolution method to learn the relationships between adjacent points, and improves the local area overlap problem that exists in the Multi-Scale Grouping(MSG) [2] module in PointNet++. These methods all regard how to aggregate local features as the main research point, but ignore the redundancy and error information existing in the local features. Thus, we propose VA-GCN which utilize the geometric relationship of point clouds to constrain the aggregation process of local information. 
\section{Method}
To accurately analyze each point in complex point clouds, not only the information of each point but also the information of its neighbor points should be taken into consideration. Inspired by DGCNN [9] and FR3DNet [10], we propose VAConv, which exploits low-dimensional geometric relations to explicitly model the attention weight matrix of edge features to highlight the more important local proximity information, and each point can perceive the surrounding point cloud structure. The VA-GCN is constructed by stacking the VAConv module of different receptive fields, which makes full use of the local information of various scales while maintaining the geometric structure in the 3D Euclidean space. Finally, max-pooling is used to aggregate local multi-scale information.
In Section 3.1, the basic structure of VAConv is introduced. In Section 3.2, the approach to stack VAConvs with different receptive fields to obtain a VA-GCN network is described.

\subsection{Vector Attention Convolution}
 
VAConv is the heart of the VA-GCN. We exploit global features $X^{w-1}=\left\{x_{1}^{w-1}, \ldots, x_{n}^{w-1}\right\} \subseteq R^{F^{n-1}}$ and absolute position of point clouds $P=\left\{p_{1}, \ldots, p_{n}\right\} \subseteq R^{3}$ as the input of VAConv module, $w-1$ is the number of layers, and $F^{w-1}$ is the number of channels output of the $w-1$ layer. As is shown in Figure \ref{fig1}, we first use the k-nearest neighbor (KNN) algorithm to select the $k$ nearest neighbor points $Q_{i}=\left\{q_{i 1}, \ldots q_{i k}\right\}$ in the range of $r$ near the center point $p_i$, and then establish a directed graph between the center point and its neighbor points to obtain the relative position vector $e_{i j}$ and edge features $D_{i j}^{w}$ of global features. By changing the size of $r$, we can flexibly adjust the receptive field. VAConv is a dual-channel structure module, the first channel exploits the relative position vector and two angles ($E_{i j}$, $A_{i j}$) of the three-dimensional space coordinate system to

\begin{figure}[htbp]
    \centering
    \includegraphics[width=\linewidth]{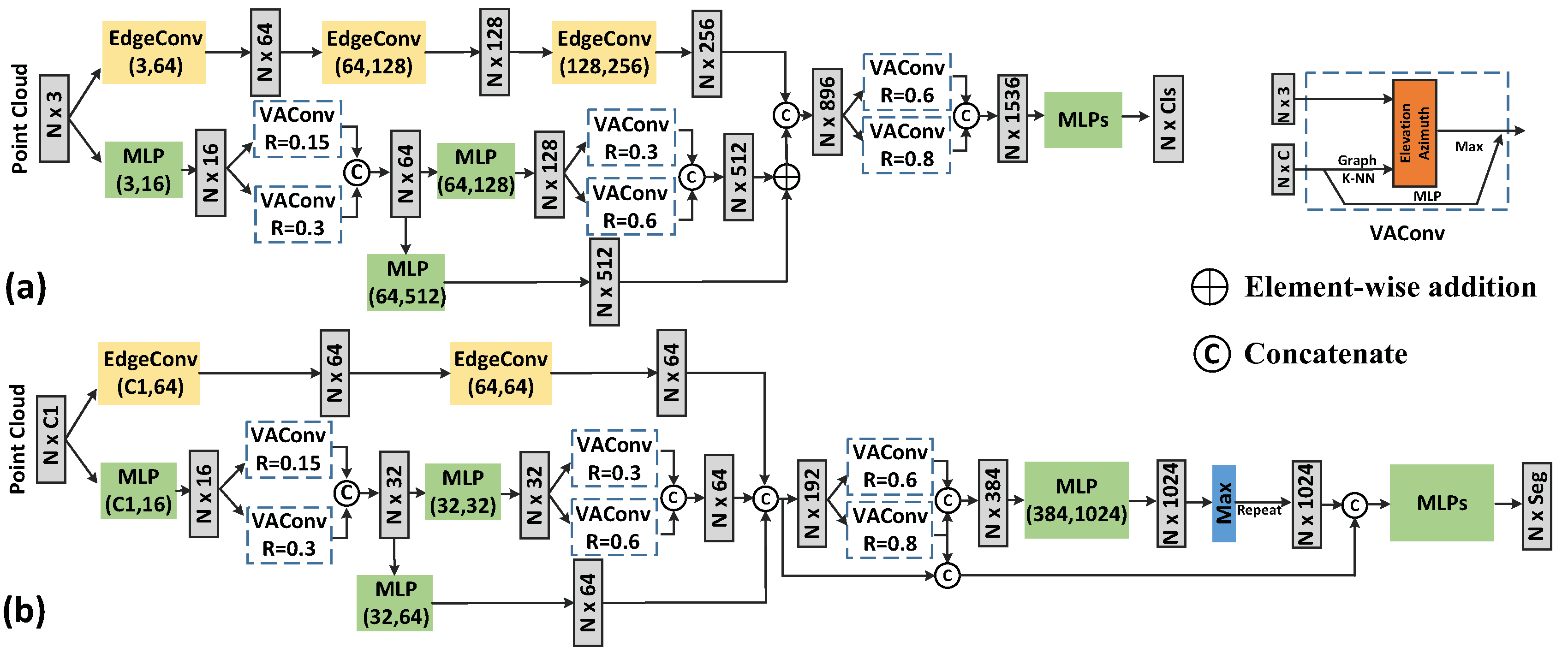}
    \caption{Schematic diagram of VA-GCN structure. The upper model is for 3D classification, and the lower is for 3D segmentation. N represents the number of input point clouds, the input of the classification network is the absolute position of the point cloud (Nx3), and the input of the segmentation network is (NxC1). The output of the classification network (NxCls) and the output of the segmentation network (NxSeg). The blue module is a schematic diagram of VAConv. Its input is the global information (NxC) extracted by the previous MLP and the absolute position of the point cloud (Nx3). After calculating the elevation and azimuth by the orange module, edge features is aggregated by max-pooling.}
    \label{fig2}
\end{figure}

explicitly construct the attention weight of edge features. The second channel uses the relative distance between the center point and its neighbor points to constrain the extraction process of high-dimensional information. A channel-wise max-pooling is used to aggregate local information to the center point.

In the first channel, we express the relative position vector $e_{i j}$ between the center point $p_i$ and its neighbor points $q_{i j}$ as:
\begin{equation}
    e_{i j}=\left\{q_{i 1}-p_{i}, \ldots, q_{i k}-p_{i}\right\}, i=\{1, \ldots, n\}
\end{equation}
Edge features $D_{i j}^{w}$ is expressed as:
\begin{equation}
   D_{i j}^{w}=\left\{x_{i 1}^{w-1}-x_{i}^{w-1}, \ldots, x_{i k}^{w-1}-x_{i}^{w-1}\right\}, i=\{1, \ldots, n\}
\end{equation}

Unlike EdgeConv [9], We did not directly concatenate the edge features into the global features. Instead, the low-dimensional geometric structure is used to constrain aggregation process of edge features, giving each edge a different weight. We take the low-dimensional geometric structure information into consideration because we believe that the original geometric structure have changed in the high-dimensional space, and simply concatenate edge features and the global information will lose the information of original geometric structure. At the same time, The distance $d i s_{i j}$ between the center point itself $p_i$ and its neighbor points $q_{i j}$ in Euclidean space can be obtained by the relative position vector $e_{i j}$:
\begin{equation}
    \left\{\begin{array}{l}
\vec{e_{i j}}=\left\{\vec{x_{i j}}, \vec{y_{i j}}, \vec{z_{i j}}\right\} \\
d i s_{i j}=\sqrt{\left(\vec{x_{i j}}\right)^{2}+\left(\vec{y_{i j}}\right)^{2}+\left(\vec{z_{i j}}\right)^{2}}
\end{array}\right.
\end{equation}

As is shown in (c) of Figure \ref{fig1}, We first establish an orthogonal three-dimensional space coordinate system XYZ, then project the relative position vector $e_{i j}$ to the XY plane. The angle between the relative position vector and the projection vector of the XY plane in the Z direction is called elevation $E_{i j}$, and the angle between the projection vector and the Y axis is called the azimuth $A_{i j}$. They can be calculated according to the position vector $e_{i j}$ and the distance $d i s_{i j}$ between adjacent points:
\begin{equation}
    \left\{\begin{array}{l}
E_{i j}=\frac{\vec{z_{i j}}}{d i s_{i j}} \\
A_{i j}=\frac{\vec{x_{i j}}}{\sqrt{\left(\vec{x_{i j}}\right)^{2}+\left(\vec{y_{i j}}\right)^{2}}}
\end{array}\right.
\end{equation}
VAConv uses MLP to extract high-dimensional edge features. $H$ is the weight function used to extract edge features, and $M(\vec{p}, \vec{q})$ is the distance of the center point to its neighbor points. Shorter the distance is, greater weight is assigned to the corresponding edge features. The purpose of the aggregation function $G(\vec{p}, \vec{q})$ is to aggregate edge features to the center point $p_i$. Their expressions are as follows:
\begin{equation}
   M(\vec{p}, \vec{q})=\frac{\left(\max \left(d i s_{i j}\right)-d i s_{i j}\right)^{2}}{\operatorname{sum}\left(\left(\max \left(d i s_{i j}\right)-d i s_{i j}\right)^{2}\right)}
\end{equation}
\begin{equation}
    G(\vec{p}, \vec{q})=\operatorname{sum}\left(H\left(D_{i j}^{w}\right) \otimes\left(\cos \left(\vec{E_{i j}}\right) \odot \cos \left(\vec{A_{i j}}\right)\right) \otimes M(\vec{p}, \vec{q})\right)
\end{equation}
In the other channel, we use relative position vector and relative distance to constrain the extraction process of global information. The first MLP is used to extract the attention weight of the global information, and after the cross-multiplication with the global information, high-dimensional feature is extracted through the second MLP.
\begin{equation}
    g^{w}=m l p\left(X^{w-1} \otimes \max \left(m l p\left(\operatorname{concat}\left(e_{i j}, d i s_{i j}\right)\right)\right)\right)
\end{equation}

After adding the outputs of the two channels, passing through the max-pooling layer which aggregate features to the center point, the final output could be achieved:
\begin{equation}
   X^{w}=\max \left(g^{w} \oplus G(\vec{p}, \vec{q})\right)
\end{equation}

\subsection{Structure of VA-GCN}

VA-GCN is a dual-channel model that obtains global shape descriptors by fusing the relative positions of spatial features in different dimensions. Its input is the original representation of point clouds, denoted as $X=\left\{x_{1}, \ldots, x_{n}\right\} \subseteq R^{C}$, when $c=3$, the input is the point cloud three-dimensional coordinates $x_{i}=\{x, y, z\}$. As is shown in Figure \ref{fig2}, in the channel (a), the input of the EdgeConv module is global information. This channel cascades three EdgeConv modules. Each EdgeConv integrates edge features into the input global information, and uses MLP to extract high-dimensional global information from it.

We stack two layers of VAConv modules to establish the channel (b). In each layer, there are two VAConv modules with different receptive fields arranged in parallel. In addition, we also gradually increase the size of the receptive fields in each layer. Firstly, we extract high-dimensional global information through MLP. Secondly, after establishing the directed graph, we exploit the relative position relationship to aggregate edge features in the VAConv module. In order to better preserve the multi-scale local information, we use MLP to directly extract the high-dimensional features of the small-scale global information and add it to the output of the second layer to obtain fused multi-scale global information.

The main function of channel (c) is to fuse the global information extracted by the two channels. We concat the output of the two channels through a third-layer VAConv module with $r=0.6$ and $0.8$, which aims to use a larger receptive field to obtain the overall geometric structure of point clouds. Finally, the point cloud global shape descriptor is obtained, which can be input into the classifier for tasks such as 3D shape recognition and 3D segmentation.
\section{Experiments}

To verify the effectiveness of the VA-GCN, we performed 3D classification, 3D semantic segmentation, and 3D part segmentation experiments on the ModelNet40 [8], S3DIS [40] and ShapeNet [35] benchmark datasets separately. For training, we use Adam optimizer under the weight decay of 0.0001, the mini-batch size is 16, and the training process starts with a learning rate of $0.001$. A cosine annealing algorithm is applied to reduce the learning rate to $0$ in $500$ epochs. We adopt PyTorch 1.1.0 framework to implement experiments on a computer with 3.4 GHz Intel Xeon-E5-2643-v3 CPU, 64G RAM, and two NVIDIA GTX 1080Ti GPUs.

\subsection{3D classification experiment and discussion}

\textbf{Dataset:}  We evaluated our model on the ModelNet40 dataset for 3D classification task. ModelNet40 has a total of
\begin{table}[htbp]
\caption{ModelNet40 shape classification results. The Overall Accuracy of VA-GCN is 2.1\% higher than that of DGCNN, and Mean Class Accuracy increased by 1.2\%. $^*$ indicates baseline. }
\label{tab1}
\centering
\resizebox{\textwidth}{!}{%
\begin{tabular}{cccccccc}
\toprule
& Method               & input          & Overall Accuracy    & Mean Class Accuracy     \\
      \toprule
      & PointNet [1]       & coordinates      &89.2\% & 86.2\%\\
      & MO-Net [29]          & coordinates     & 89.3\% & 86.1\%\\
      & Deep Sets [30]       & coordinates      & 90.3\% & -\\
      & PointNet++ [2]       & coordinates      & 90.7\% & -\\
      & PointCNN [3]       & coordinates      & 92.2\% & 88.1\%\\
      & PCNN [31]       & coordinates      & 92.3\% & -\\
      & A-CNN [27]      & coordinates      & 92.6\% & 90.3\%\\
      & Point2Seq [32]       & coordinates      & 92.6\% & -\\
      & KPConv [19]      & coordinates      & 92.7\% & -\\
      & PointASNL [22]       & coordinates      & 92.9\% & -\\
      & PointASNL [22]      & coordinates+normal      & 93.2\% &-\\
      & Geo-CNN [26]      & coordinates+normal      & 93.4\% & 91.1\%\\
      \bottomrule
      & DGCNN$^*$ [9]      & coordinates      & 92.2\% & 90.2\%\\
      & VA-GCN      & coordinates      & 92.7\%  $\textcolor{red}{\uparrow}$0.5\% & 89.3\%\\
      & VA-GCN       & coordinates+normal      & 93.5\%  $\textcolor{red}{\uparrow}  $1.3\% & 90.4\%  $\textcolor{red}{\uparrow}$0.2\%\\
      & \textbf{VA-GCN+MSI}   & coordinates+normal 
      & \textbf{94.3\%} $\textcolor{red}{\uparrow}$2.1\%
      &\textbf{91.4\%} $\textcolor{red}{\uparrow}$1.2\%\\
\bottomrule
\end{tabular}}
\end{table}
12311 CAD models, among which there are 9843 training samples and 2468 test samples. For each training sample, we uniformly sampled 1024 points. We use two sets of data for training, one with the normal and the other without. During the training procedure, we augment the data [2] by scaling objects and perturbing the object and point locations. Overall accuracy and mean class accuracy are adopted as indicators of the performance.

\textbf{Implementation Details:}  On the ModelNet40 dataset for 3D classification, the parameters and structure of our model are shown in the classification network of Figure \ref{fig2}. The last MLPs layer contains 3 fully connected layer, the sizes of them are (2048,512), (512,256), (256,40) separately. 

\textbf{Result:}  As is shown in Table \ref{tab1}, the VA-GCN achieved the most advanced performance on the ModelNet40 dataset for 3D classification task. Inspired by multi-scale inference [1,2], we use multi-sample inference(MSI) for inference. That is, we sampled 1024 points out of input points randomly for inference, and repeated this procedure for 10 times, then averaged the 10 predicted results as the final result. This method can further improve the overall accuracy by 0.8\%.

\subsection{3D semantic segmentation experiment and discussion}

\textbf{Dataset:}  We evaluated our model on S3DIS [36] dataset for semantic scene segmentation task. S3DIS consists 272 point cloud scanning models of 6 indoor areas. Each point in the model belongs to one of the 13 pre-classifications. We divided the room into blocks with an area of 1m$^{2}$, using XYZ, RGB and normalized location to represent point clouds. We sampled 4096 points from each model for training, and utilized Area-5 as the test scene. 
\textbf{Implementation Details:} For semantic segmentation task, the parameters and structure of our model are shown in the segmentation model of 

\begin{figure}[htbp]
    \centering
    \includegraphics[width=\linewidth]{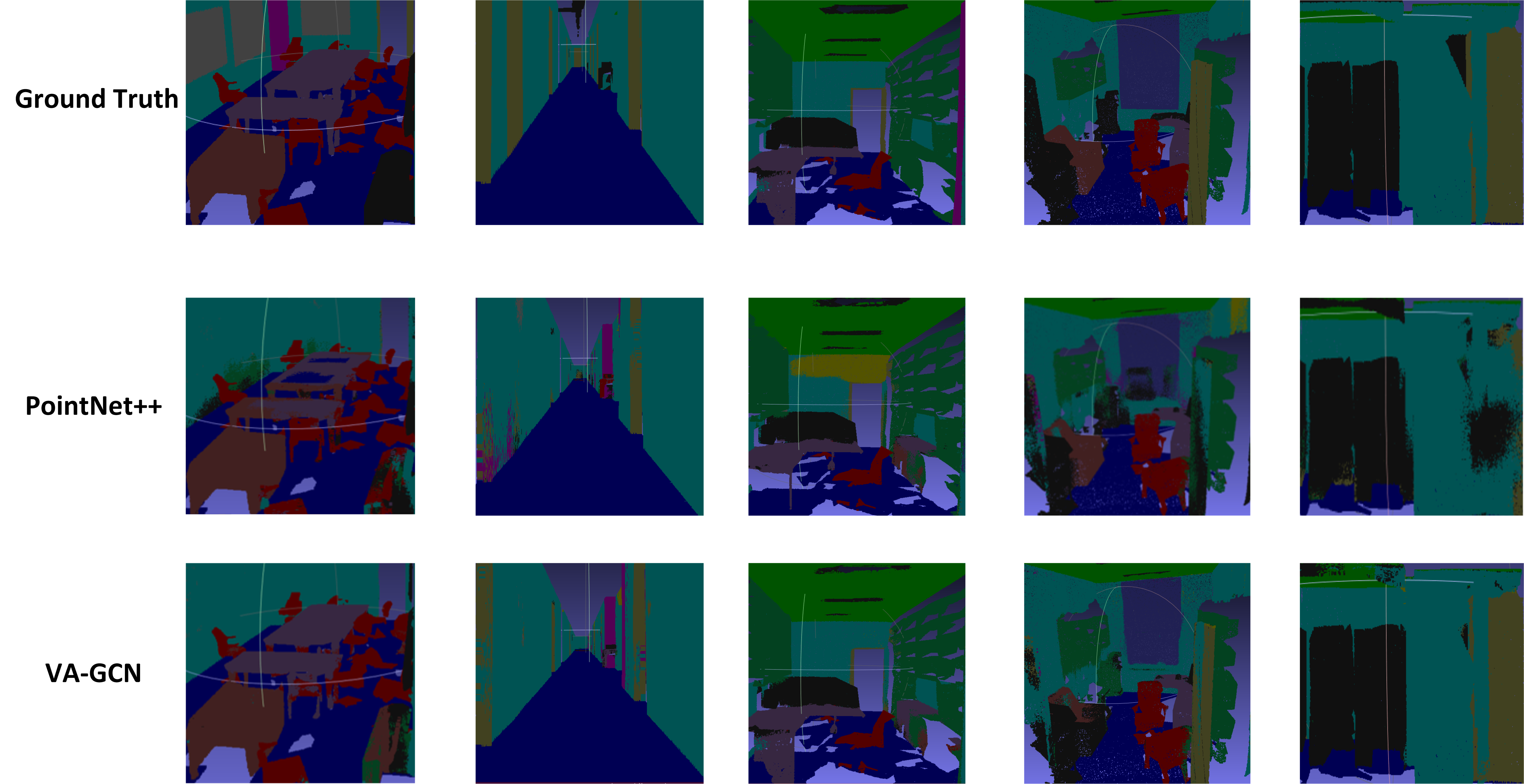}
    \caption{visualization results of S3DIS, from top to bottom are ground truth, PointNet++ and VA-GCN separately.}
    \label{fig4}
\end{figure}

Figure \ref{fig2}. The input of the network is XYZ, RGB and standardized spatial coordinates (Nx9). The last MLPs layer contains 3 fully connected layer, the sizes of them are(1408,512), (512,256), (256,13). 

\textbf{Result:}  As is shown in Table \ref{tab2}, We use mean Inter-over-Union (mIoU) as evaluation indicator. Our VA-GCN results are 0.8\% higher than the result of DGCNN with 6-fold cross validation(calculating the metrics with results from different folds merged). Figure \ref{fig4} shows the visualization results on the S3DIS dataset, in visualization, our results are significantly better than PointNet++. 

\begin{table}[htbp]
\caption{Results of S3DIS indoor semantic segmentation on Area-5. The metrics of PointNet++ are generated by publicly available code. $^*$ indicates baseline and DGCNN use 6-fold cross validation. }
\label{tab2}
\centering
\resizebox{\textwidth}{!}{%
\begin{tabular}{ccccccccc}
\toprule
&Method &PointNet [1]  & SegCloud [37]   &RSNet [28]  &PointNet++ [2] &DGCNN$^*$ [2] &\textbf{VA-GCN}\\
      \toprule
      & mIoU  & 41.1\%  & 48.9\%  &51.9\%   &55.0\% &56.1\%  &\textbf{56.9}\%$\textcolor{red}{\uparrow}$0.8\%\\
\bottomrule
\end{tabular}}
\end{table}

\subsection{3D part segmentation experiment and discussion}

\textbf{Dataset:}  We conducted 3D part segmentation experiments on the ShapeNet dataset. The dataset contains 16881 models divided into 16 categories, with a total of 50 parts marked. Each point in the point clouds collection is marked as one of the preset categories. We sampled 2048 points from each model for training. Afterwards, we performed voting tests with random scaling and then averaged the prediction results [1,2]. 

\textbf{Implementation Details:} For the 3D part segmentation task, the specific structure of our VA-GCN is shown in Figure \ref{fig2}(b). The input of the network are XYZ (Nx3) and a one-hot label (1x16). The one-hot label passes through the MLP with the size of (16,64) and then is merged with the global features. The last MLPs layer contains 3 MLP, the sizes of them are (1472,512), (512,256), (256,13) separately. We used the instance average and class average mean Inter-over-Union (mIoU) as the evaluation index. 

\begin{table}[htbp]
\caption{Metrics results on ShapeNet for 3D part segmentation task. $^*$ indicates baseline. }
\label{tab3}
\centering
\resizebox{\textwidth}{!}{%
\begin{tabular}{cccccccccc}
\toprule
&Method &PointNet [1]  & PointNet++ [2]   &3D-GCN [39] & PCNN [31] &DGCNN$^*$ [9]&\textbf{VA-GCN}\\
      \toprule
      & Cls. mIoU  & 80.4\%  & 81.9\% &82.1\% &81.8\%  &82.3\% &\textbf{82.6}\%$\textcolor{red}{\uparrow}$0.3\% \\
      & Ins. mIoU  & 83.7\%  & 85.1\% &85.1\%  &85.2\%  &85.2\% &\textbf{85.5}\%$\textcolor{red}{\uparrow}$0.3\% \\
\bottomrule
\end{tabular}}
\end{table}

\begin{figure}[htbp]
    \centering
    \includegraphics[width=\linewidth]{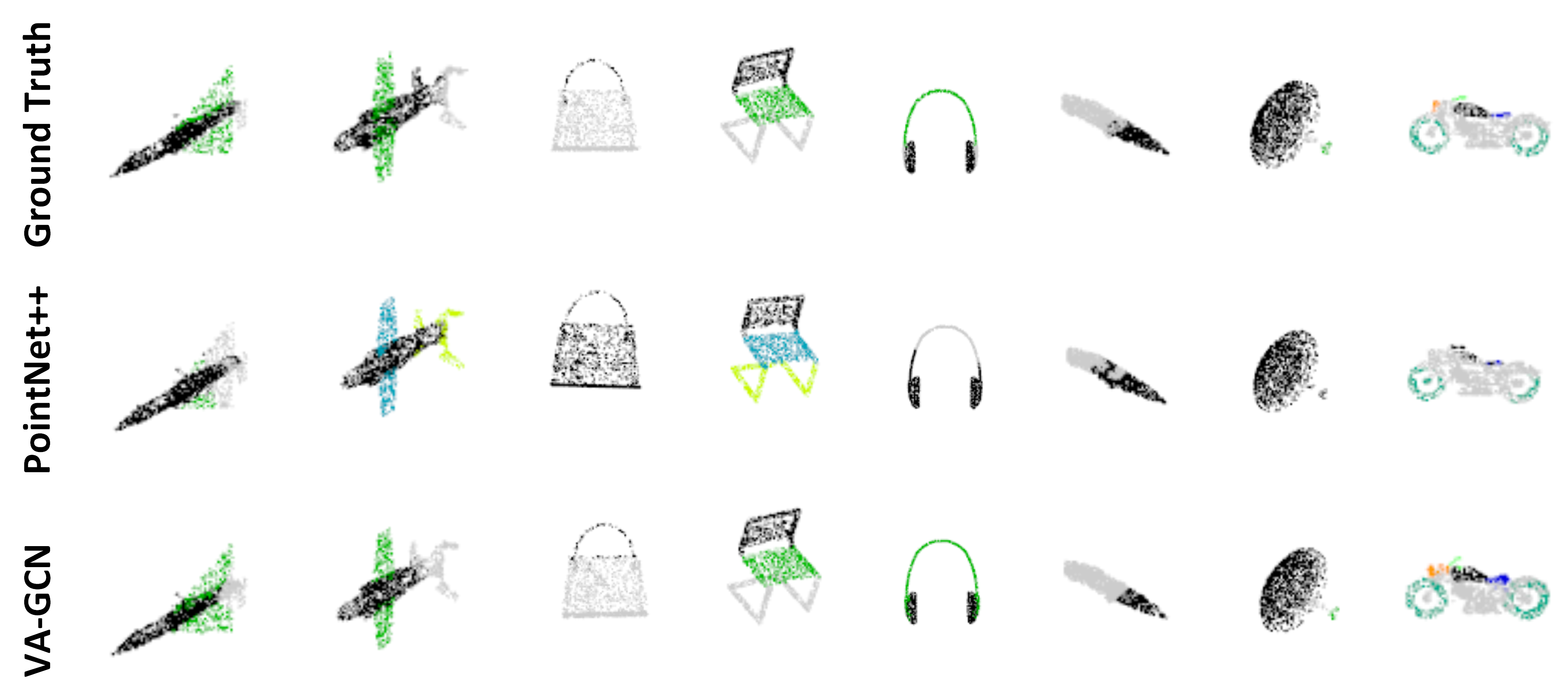}
    \caption{Visualization results on ShapeNet for part segmentation task. The first column is the ground truth, the second column is the results of PointNet++, and the third column is the results of VA-GCN. From left to right are rocket, airplane, bag, chair, earphone, knife, lamp, motorbike. Same color is uesd to label the same class. }
    \label{fig5}
\end{figure}

\textbf{Result:}  Table \ref{tab3} lists the instance average and class average
mean Inter-over-Union (mIoU). The Cls.mIoU and the Ins.mIoU of VA-GCN is 0.3\% better than DGCNN. Figure \ref{fig5} is the visualized 
segmentation results, compared with PointNet++, our VA-GCN has more detailed segmentation and can better classify the part of the object.

\subsection{Ablation experiments}

In the ablation experiments, we verified the effectiveness of the parallel structure of VAConv modules with different scales. As shown in Table \ref{tab4}, we build four VA-GCN models with different numbers of parallel structure: v0 does not use parallel structure; v1 uses one parallel structure in the first layer;v2 uses parallel structure in the first two layers ; v3 uses one parallel structure in each layer of VA-GCN. We doubled the output channels of the VAConv module in a single structure to make it identical with the number of output channels of the parallel structure. The ablation experiment proved the effectiveness of the parallel VAConv structure. As is shown in Table \ref{tab4}, the v3 model obtained the best performance in the classification task, and Overall Accuracy improved by 0.9\% compared with not using the parallel structure.

\begin{table}[htbp]
\caption{Comparison of the Overall Accuracy of models with different parallel channel numbers. }
\label{tab4}
\centering
{
\begin{tabular}{cccccccccc}
\toprule
&Method &VA-GCNv0  & VA-GCNv1  &VA-GCNv2 &\textbf{VA-GCNv3}\\
      \toprule
      & Overall Accuracy  & 92.6\%  & 93.3\%  &92.7\% &\textbf{93.5\%}\\
\bottomrule
\end{tabular}}
\end{table}

As is shown in Table \ref{tab5}, we compared performance of the single-channel structure and the dual-channel structure in the ModelNet40 classification task. The Overall Accuracy of the dual-channel structure is 0.7\% higher than that of the single-channel structure. This proves that the fusion of high-dimensional edge features and low-dimensional geometric information helps the point cloud analysis model to better capture the information of points.

\begin{table}[htbp]
\caption{Comparison of the Overall Accuracy of models with different parallel channel numbers. }
\label{tab5}
\centering
{
\begin{tabular}{cccccccccc}
\toprule
&Method & Only EdgeConv channel [9]  & Only VAConv channel  &\textbf{Dual channel}\\
      \toprule
      & Overall Accuracy  & 83.1\%  & 92.8\%  & \textbf{93.5\%}\\
\bottomrule
\end{tabular}}
\end{table}

\section{Conclusion}
In this article, we propose a new dual-channel multi-scale aggregation model VA-GCN, the core of which is the VAConv module. Owing to the stack of VAConv with different receptive fields and EdgeConv, the global features are fully integrated. The VAConv makes use of the relative position relationship of the point cloud neighborhood to calculate the elevation and the azimuth, then utilizes these two angles to explicitly model the attention weight matrix for edge features. VA-GCN achieves state-of-the-art performance on the challenging ModelNet40, S3DIS and ShapeNet datasets. To the best of our knowledge, the proposed VA-GCN achieves the highest score on ModelNet40 dataset. 

\end{document}